\documentclass[a4paper,11pt]{article}

\usepackage[hidelinks]{hyperref}
\usepackage{graphicx}
\usepackage{amsmath}
\usepackage{amsfonts}
\usepackage{parskip}
\usepackage{subcaption}
\usepackage{url} 
\usepackage{booktabs} 
\usepackage{times}
\usepackage{natbib}
\usepackage{enumitem}
\usepackage[margin=25mm]{geometry}

\title{Using Multi-Sense Vector Embeddings for Reverse Dictionaries}
\date{}
\author{
  Michael A. Hedderich\textsuperscript{1}, Andrew Yates\textsuperscript{2}, Dietrich Klakow\textsuperscript{1} and Gerard de Melo\textsuperscript{3} \vspace{0.2cm}\\ 
  \textsuperscript{1}Spoken Language Systems (LSV), Saarland Informatics Campus, Saarbr\"ucken, Germany\\
  \textsuperscript{2}Max Planck Institute for Informatics, Saarland Informatics Campus, Germany \\
  \textsuperscript{3}Rutgers University, New Brunswick, NJ, USA \\
  \texttt{\{mhedderich, dietrich.klakow\}@lsv.uni-saarland.de,}\\
  \texttt{ayates@mpi-inf.mpg.de, gdm@demelo.org}
}

\newcommand{\es}{e_\mathrm{s}}
\newcommand{\emulti}{e_\mathrm{m}}
\newcommand{\realnumbers}{\ensuremath{\mathbb{R}}}
\renewcommand{\vec}[1]{{\bf{#1}}}

\begin{document}

\maketitle
\thispagestyle{empty}
\pagestyle{empty}

\begin{abstract}
Popular word embedding methods such as word2vec and GloVe assign a single vector representation to each word, even if a word has multiple distinct meanings. Multi-sense embeddings instead provide different vectors for each sense of a word. However, they typically cannot serve as a drop-in replacement for conventional single-sense embeddings, because the correct sense vector needs to be selected for each word. In this work, we study the effect of multi-sense embeddings on the task of reverse dictionaries. We propose a technique to easily integrate them into an existing neural network architecture using an attention mechanism. Our experiments demonstrate that large improvements can be obtained when employing multi-sense embeddings both in the input sequence as well as for the target representation. An analysis of the sense distributions and of the learned attention is provided as well.
\end{abstract}

\section{Introduction}

\begin{figure}
    \begin{subfigure}{0.5\textwidth}
        \centering
        \includegraphics[width=0.8\textwidth,trim={1.9cm 1cm 1.5cm 1cm},clip]{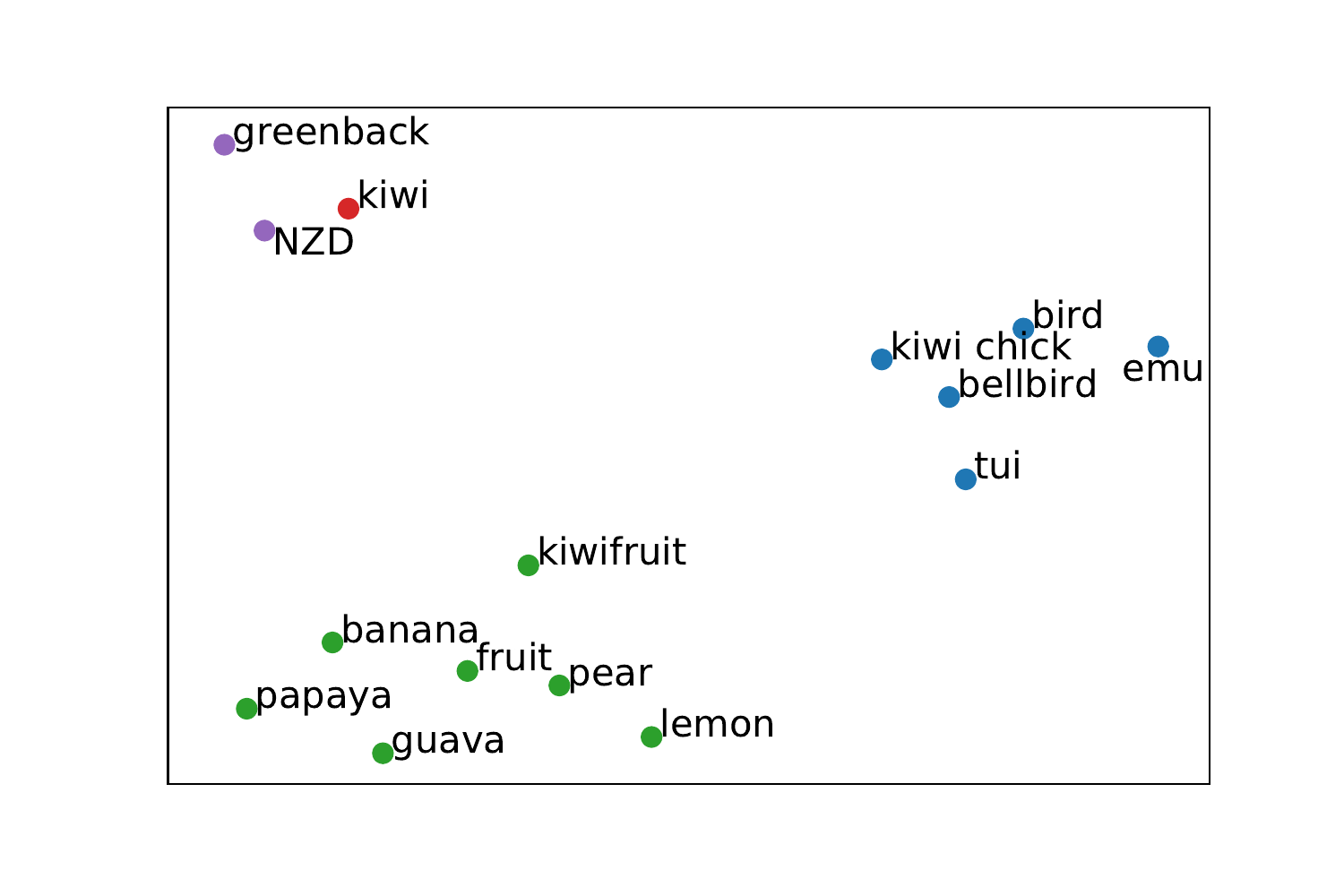}
        \subcaption{}
    \end{subfigure}
    \begin{subfigure}{0.5\textwidth}
       \centering
       \includegraphics[width=0.8\textwidth,trim={1.9cm 1cm 1.5cm 1cm},clip]{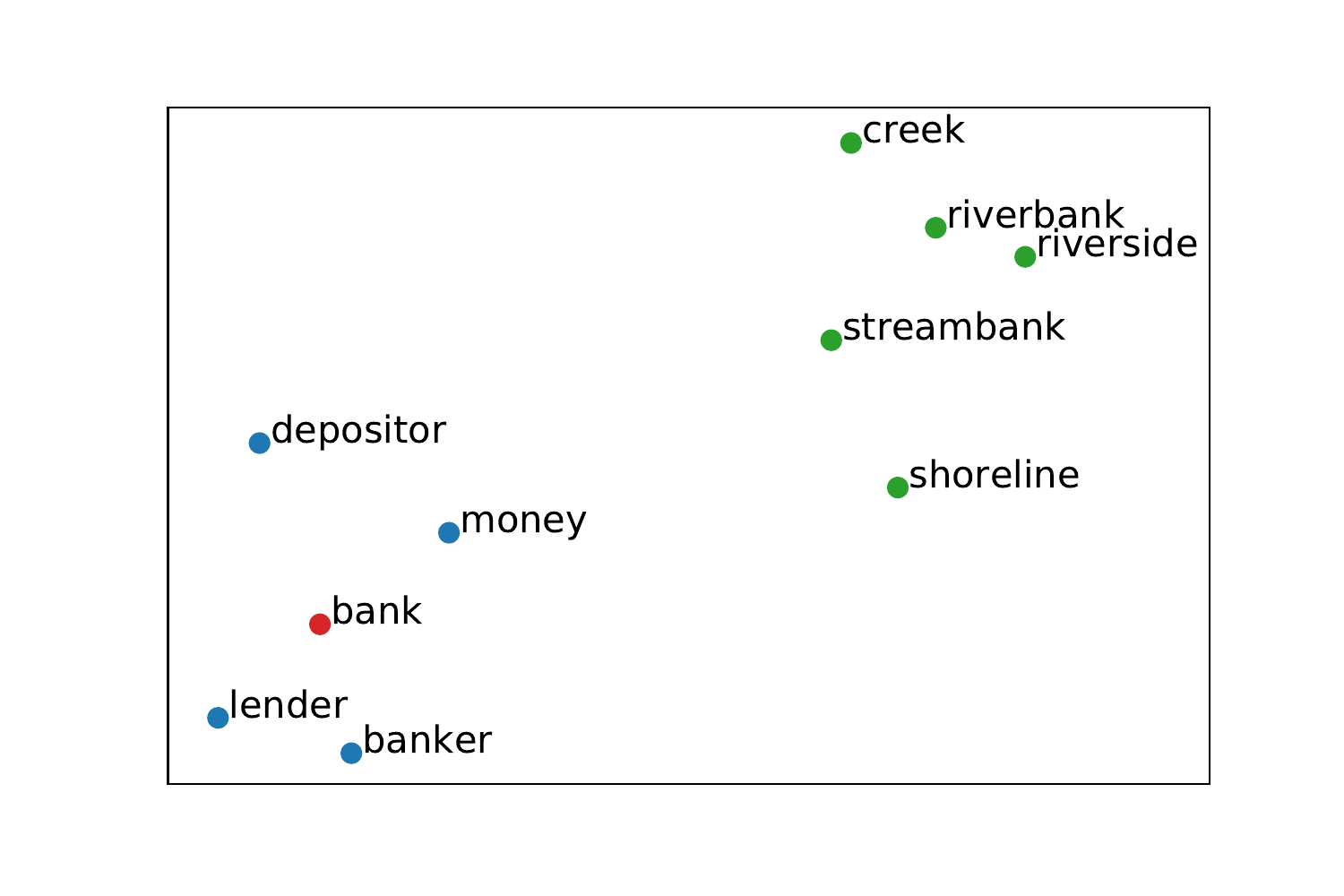}
       \subcaption{\label{fig:bank-river-vs-financial}}
     \end{subfigure}
    \caption{2D projections of Google News word2vec vectors using t-SNE \citep{maaten2008visualizing}. 
    The vector for the word \textit{kiwi} is located near the embedding for the New Zealand dollar (violet) and not near other birds (blue) or fruits (green). For \textit{bank}, the vector lies in a neighborhood of financial terms (blue), further apart from other river related terms (green). \label{fig:w2v_example_visualization} }
\end{figure}

One problem with popular word embedding methods such as word2vec \citep{word2vec} and GloVe \citep{pennington2014glove} is that they assign polysemic or homonymic words the same vector representation, i.e., words that share the same spelling but have different meanings obtain the same representation. For example, the word ``\emph{kiwi}" can signify either a green fruit, a bird or, in informal contexts, the New Zealand dollar, which are three semantically distinct concepts. If only a single vector representation is used, then this representation is likely to primarily reflect the word's most prominent sense, while neglecting other meanings (see Figure \ref{fig:w2v_example_visualization}). More generally, a word vector may be a linear superposition of features of multiple unrelated meanings \citep{Arora2018LinearAlgebraicStructure}, resulting in incoherent vector spaces. 

In recent years, several ideas have been proposed to overcome this problem. They have in common that they obtain different vector representations for the different meanings of polysemes or homonyms. Most prior work only evaluates these multi-sense vectors on single word benchmarks, however, and there is comparably little evidence for the benefits of using these embeddings in other applications. 

One multi-word task that suffers from the presence of polysemy and homonymy is the building of a reverse dictionary that can take definitions of words as input and infers the corresponding words. In this work, we present the following contributions: (1) We show that multi-sense vectors are a better representation for the target words in this task. (2) We propose a technique to select multi-sense vector embeddings for the words in the input sequence. It is based on an attention mechanism and can be incorporated into an existing end-to-end neural network architecture outperforming single-sense vector representations. (3) We provide a comparison between pre-trained and task-specific multi-sense embeddings as well as contextual word embeddings. (4) We analyze the distribution of multi-sense words in the data and the attention the network learns.
    
\section{Task and Architecture}

In this section, we give further details on the different embeddings, the reverse dictionary task and the corresponding architecture. We also motivate the use of multi-sense embeddings for the target and input vectors with qualitative examples and a quantitative analysis. 

\subsection{Single- and Multi-Sense Word Embeddings}

A single-sense word embedding $\es$ maps a word or token to an $l$-dimensional vector representation, i.e.\ $\es(w_i) = \vec{d}_i \in \realnumbers^l$ for a word $w_i$. They are often pre-trained on large amounts of unlabeled text and serve as a fundamental building block in many neural NLP models. Popular word embeddings include word2vec, GloVe and fastText \citep{bojanowski2017enriching}. If a word has several meanings, these are still mapped to just a single vector representation.

\begin{figure}
    \centering
    \includegraphics[width=0.4\textwidth]{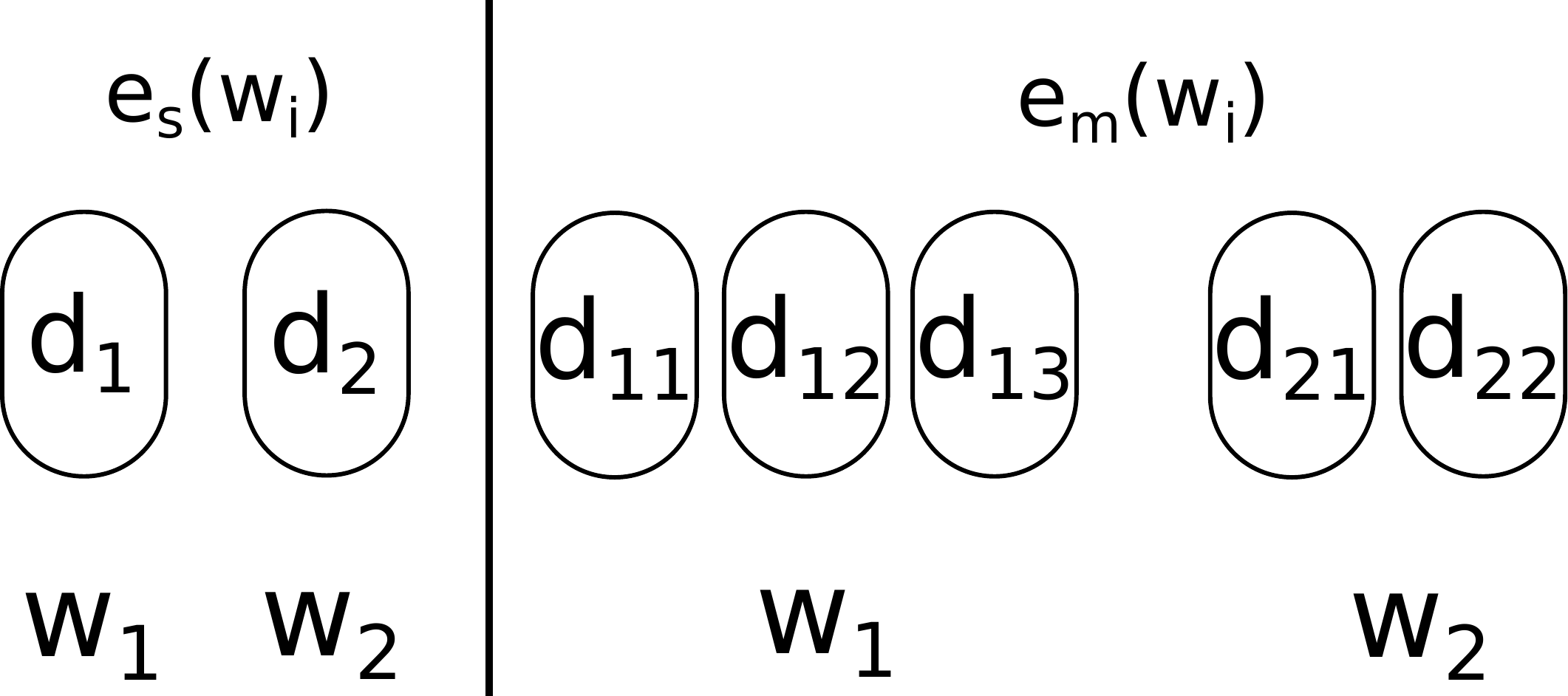}
    \caption{Single-sense embedding $\es$ compared to multi-sense embedding $\emulti$ for a sequence of input words $w_1$, $w_2$. \label{fig:single-vs-multi}}
\end{figure}

Multi-sense word embeddings $\emulti$ overcome this limitation by mapping each word $w_i$ to a list of sense vectors $\emulti(w_i) = (\vec{d}_{i1}, ..., \vec{d}_{ik})$, where $k$ is the number of senses that one considers $w_i$ to have. The vector $\vec{d}_{ij}$ then represents one sense of the given word. This difference is visualized in Figure \ref{fig:single-vs-multi}. Often, these embeddings can also be pre-trained on unlabeled text. A discussion of different multi-sense word embeddings is given in Section \ref{sec:related_work}. 

\subsection{Reverse Dictionaries}
\label{sec:reverse_dictionary}
    
A reverse dictionary is a tool for authors and writers seeking a word that is \textit{on the tip of their tongue}. Given a user-provided definition or description, a reverse dictionary attempts to return the corresponding word \citep{ZockReverseDictionaries}. We create a dataset for this task using the WordNet resource \citep{wordnet}. For each word sense in this lexical database, we consider the provided gloss description as the input, and the word as the target.

\begin{quote}
\textit{the size of something as given by the distance around it} $\rightarrow$ \textit{circumference}
\end{quote}    
    
More details about the dataset are given in Section \ref{sec:data}. \citet{hill2016dicteval} presented a neural network approach for this task and also set it in the wider context of sequence embeddings. Each instance consists of a description, i.e.\ a sequence of words $(w_1, ..., w_n)$, and a target vector $\vec{t}$. Each word of the input sequence $w_i$ is mapped with a single-sense word embedding function $\es$ (e.g.\ word2vec) to a vector representation $\es(w_i)$. This sequence of vectors is then transformed into a single vector 
    
\begin{align}
    \vec{\hat{t}} = f(\es(w_1), ..., \es(w_n)). \label{eq:hatf_original}
\end{align}
    
For $f$, the authors use---among others---a combination of an LSTM \citep{lstm} and a dense layer. The network is trained with the cosine loss between $\vec{t}$ and $\vec{\hat{t}}$. During testing or when employed by a user, the model produces a ranking of the vocabulary words $(\alpha_1, ..., \alpha_{|V|})$ by comparing the vector representation $\es(\alpha_i)$ of each vocabulary word $\alpha_i$ with the prediction $\vec{\hat{t}}$ in terms of the cosine similarity measure. The $k$ most similar words are returned to the user. We choose this task and architecture to show which benefits multi-sense vectors can bring to a downstream application and how they can easily be incorporated into an existing architecture. Two major limitations of single-sense vectors in this approach are presented in the following two subsections.
    
\subsection{Target Vectors}
\label{sec:target_vectors}

The first limitation is that of the target vector, as exposed in Figure \ref{fig:bank-river-vs-financial}. For the single-sense embedding, the vector for \textit{bank} lies in a neighborhood consisting of financial terms with words such as \textit{banker}, \textit{lender} and \textit{money}. Given a description of a river \textit{bank} as input (\textit{the slope beside a body of water}), a model trained on single-sense vectors as targets would have to produce a vector $\vec{t}$ (red point) that resides in a region of the semantic space that relates to financial institutions (blue points), rather than to nature and rivers (green points) with terms such as \textit{riverside} or \textit{streambank}. 

\begin{figure}
    \begin{subfigure}{0.5\textwidth}
        \includegraphics[width=\textwidth]{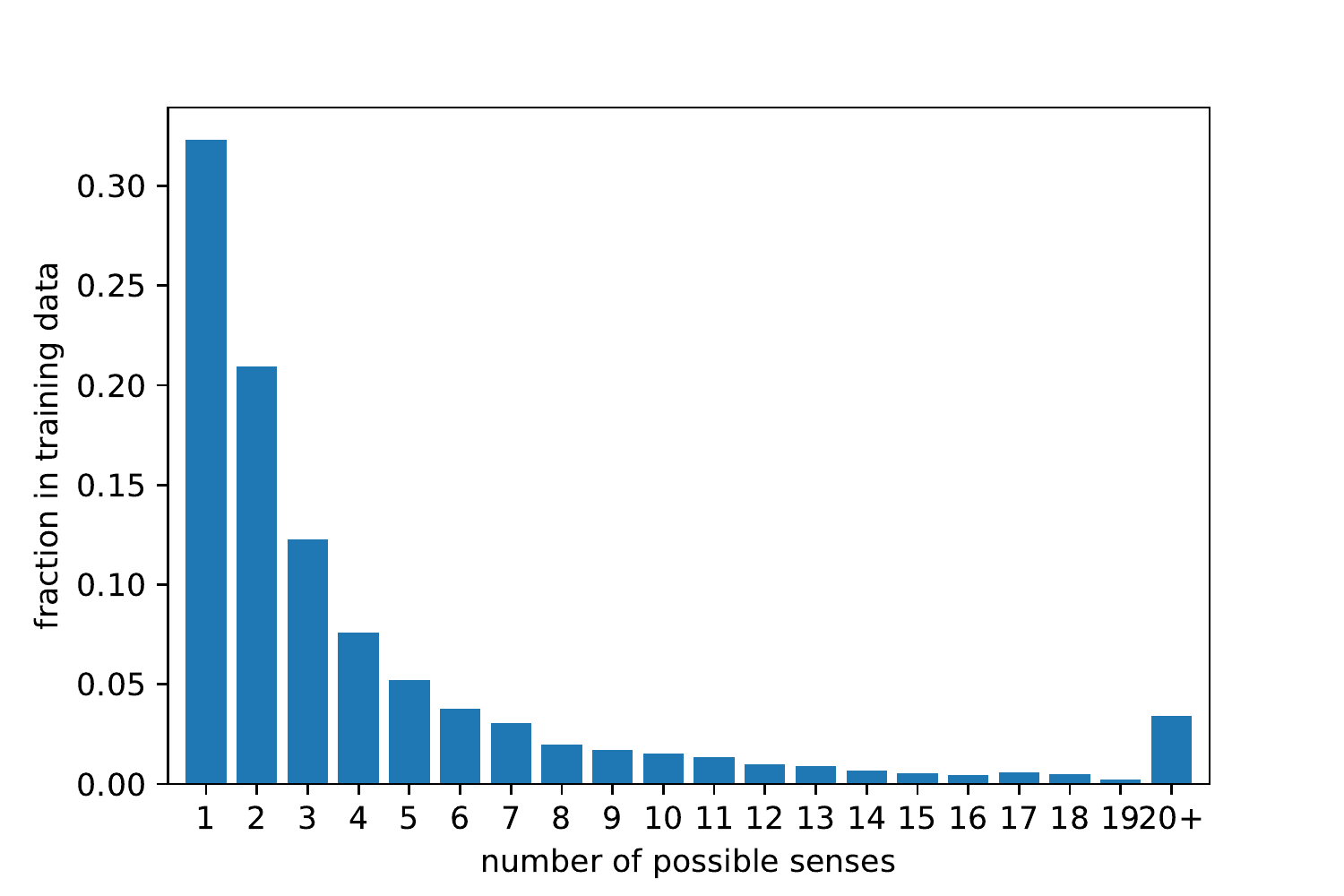}
        \caption{\label{fig:num-senses-target} Target words}
    \end{subfigure}
    \begin{subfigure}{0.5\textwidth}
        \includegraphics[width=\textwidth]{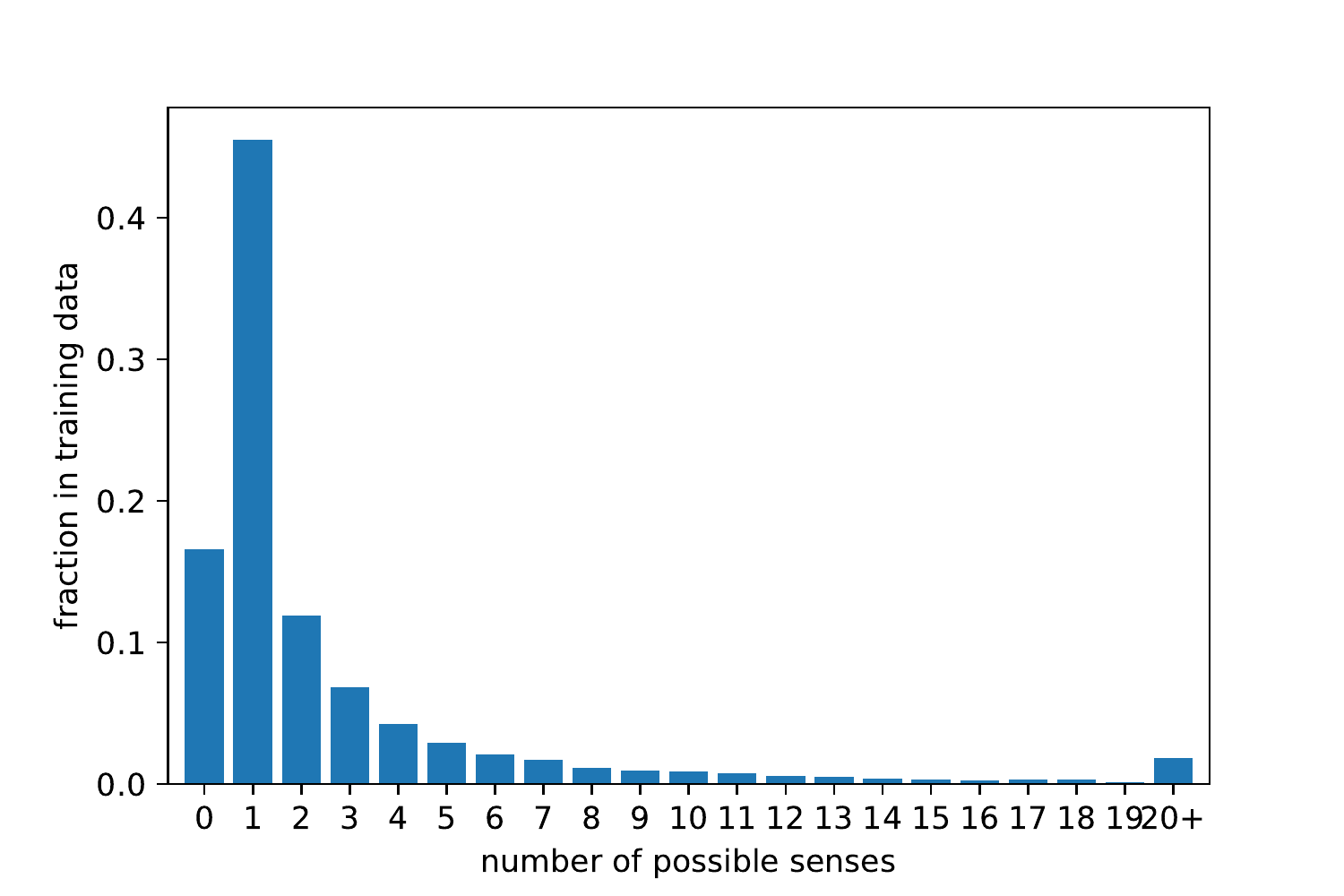}
        \caption{\label{fig:num-senses-description} Description words}
    \end{subfigure}    
    \caption{Number of possible senses, according to WordNet (see Section \ref{sec:embeddings}), of the target words (left) and input words (rights) in the training data. Out-of-vocabulary words are listed as having 0 senses.}
\end{figure}

In Figure \ref{fig:num-senses-target}, we observe that 68\% of the target words in our training data have more than one possible sense in WordNet. While the sense distinctions in WordNet tend to be rather fine-grained, this shows that in general the phenomenon of encountering multiple senses for a target word is not limited to only a few instances but affects a large portion of the data.
    
To cope with this, we propose to rely on multi-sense vectors for the target $\vec{t}$. Using these, we can assign the vector corresponding to the correct sense to each target in the training data. During testing, the correct target sense should obviously not be known to the model. We hence use for the ranking a vocabulary that consists of all sense vectors of all words. 
    
\subsection{Input Vectors}
\label{sec:input_vectors}

The second limitation of the existing architecture is the fact that it uses single-sense vectors for the input sequence. For example, within the definition of a \textit{bluff}, \textit{a high steep bank usually formed by river erosion}, the word \textit{bank} refers to the phenomenon in nature. Therefore, the vector embedding for \textit{bank} should also semantically reflect this and should not reside in a semantic region relating to the dominating, financial meaning. 

The analysis in Figure \ref{fig:num-senses-description} shows that 38\% of the words in the input sequence have more than one possible sense. This is a smaller percentage than in the case of target vectors, mostly due to out-of-vocabulary words and frequently occurring single-sense words such as stopwords. Nevertheless, this shows that multi-sense considerations are relevant for over a third of the words in the input definitions.
    
In contrast to the target vectors, we cannot directly link each input word to the correct sense vector because annotating every description word with the corresponding sense would be very expensive. Instead, we propose to provide the model with all possible sense vectors for each description input word and to perform the selection directly within the neural network architecture in an end-to-end fashion. Our approach to achieve this in a differentiable way, employing an attention mechanism, is given in the next section.
    
\section{Multi-Sense Vector Selection}
\label{sec:multi_sense_vector_selection}
    
\begin{figure}
    \centering
    \includegraphics[width=0.5\textwidth]{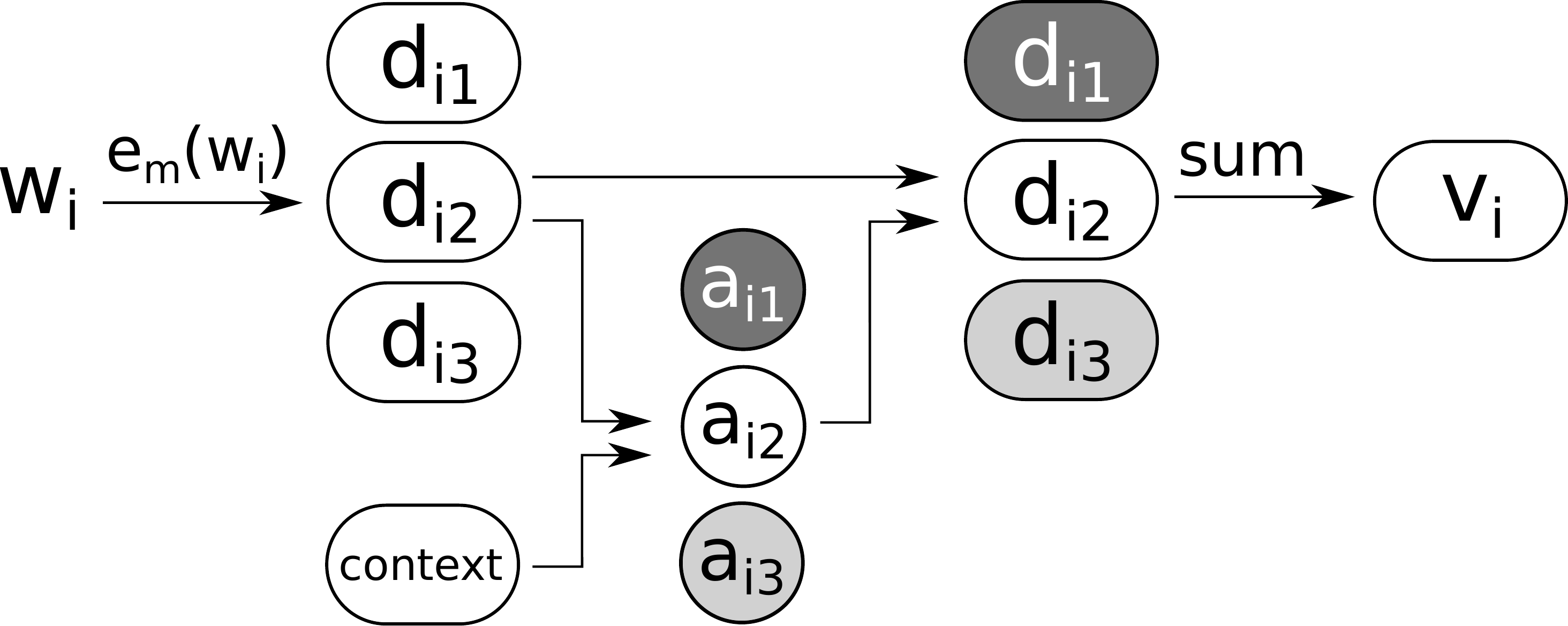}
    \caption{Visualization of the multi-sense vector selection using attention.\label{fig:multi-sense-attention}}
\end{figure}
    
The process of selecting multi-sense vectors is visualized in Figure \ref{fig:multi-sense-attention}. For an input sequence of words $(w_1,...,w_n)$, first a representation of the context is computed. For this, a single-sense word embedding function $\es$ is used and an LSTM transforms this sequence into a context vector $\vec{c}$:
    
\begin{align*}
    \vec{c} = \text{LSTM}(\es(w_1), ..., \es(w_n))
\end{align*}
    
For each word $w_i$, the multi-sense embedding function $\emulti$ provides one or more sense vectors $\emulti(w_i) = (\vec{d}_{i1}, ..., \vec{d}_{ik})$. Each sense vector $\vec{d}_{ij}$ is compared to the context by computing the raw attention
    
\begin{align}
    r_{ij} = f(\sigma(\vec{c},\vec{d}_{ij})),
\end{align}
    
where $\sigma$ is a similarity function (dot product or cosine similarity in our case) and $f$ is a non-linear function (ReLU in our experiments). The raw attention is normalized to yield attention weights
    
\begin{align}
    a_{ij} = \frac{\exp(r_{ij})}{\sum_h \exp(r_{ih})},
\end{align}
    
and we obtain a new representation
    
\begin{align}
    \vec{v}_i = \sum_{j=1}^k a_{ij} \vec{d}_{ij}.
\end{align}
    
For each input word, instead of $\es(w_i)$, the vector $\vec{v}_i$ is used in the task architecture. Equation \ref{eq:hatf_original} then becomes
    
\begin{align}
    \vec{\hat{t}} = f(\vec{v}_1, ..., \vec{v}_n).
\end{align}

\section{Experimental Evaluation}

In the following, we will detail our experiments to evaluate the effect of multi-sense embeddings both for the input description and for the target words.

\subsection{Data}
\label{sec:data}
The dataset was created by extracting all single word lemmas from WordNet version 3.0\footnote{We do not use the original dataset by \citet{hill2016dicteval} as it contains a flaw where a substantial part of the "unseen" test instances are also part of the training data.}. Each instance consists of a lemma as the target word and its corresponding definition as the description. We make this dataset publicly available\footnote{\url{https://github.com/uds-lsv/Multi-Sense-Embeddings-Reverse-Dictionaries}}. When creating the data, we used an 80\%/10\%/10\% train/dev/test split of the WordNet synsets. The data was split along synsets and not words to avoid any leakage of information from the test to the training data. For a fairer comparison with the single-sense baseline, we only used instances where the target word was in the vocabulary of the single-sense embedding. This resulted in 85,136 train, 10,521 development and 10,502 test instances. The descriptions were tokenized using SpaCy version 2.0.11 \citep{spacy2}. The distribution of the part-of-speech tags of the target words is given in Table \ref{tab:distribution-pos}.

\begin{table}
    \centering
    \begin{tabular}{lc|c|c|c}
        \toprule
        \textbf{} & \textbf{noun} & \textbf{verb} & \textbf{adj} & \textbf{adv} \\ \midrule
        target words by POS & \multicolumn{1}{c}{59\%} & \multicolumn{1}{c}{17\%} & \multicolumn{1}{c}{20\%} & 4\% \\ \midrule
        target word with 1 sense & 38\% & 6\% & 34\% & 54\%\\ 
        target word with 2 senses & 21\% & 13\% & 26\% & 21\% \\ 
        target word with 3+ senses & 41\% & 81\% & 40\% & 25\%\\
        \bottomrule
    \end{tabular}
    \caption{The first row shows the distribution of the part-of-speech tags (POS) of the target words in the dataset. The rest of the table contains the distribution of the number of senses, according to WordNet, given a specific POS. \label{tab:distribution-pos}}
\end{table}

\subsection{Embeddings}
\label{sec:embeddings}

In this work, we consider as our single-sense embedding $\es$ the popular 300-dimensional word2vec vectors trained on the Google News corpus\footnote{\url{https://code.google.com/archive/p/word2vec/}}. For the multi-sense embedding $\emulti$, we chose the DeConf embeddings by \citet{pilehvar2016conflated}, which reside in the same space 
as the word2vec embeddings.
It should be noted that DeConf uses the WordNet \textit{graph structure} for the pre-training of the embeddings, while for our reverse dictionary data we only use the WordNet \textit{glosses} as definitions.

\subsection{Baselines}
We compare our \textbf{multi-sense} approach that we introduced in the previous sections to the following baselines:

\begin{itemize}[leftmargin=*]
\item For the \textbf{single-sense} baseline, we use the reverse dictionary architecture proposed by \citet{hill2016dicteval}, which also serves as the foundation of all the multi-sense models.
\item In \textbf{first multi-sense}, we experiment with using the first multi-sense vector for every word as a single-sense vector, i.e.\ $\vec{v}_i = \vec{d}_{i1}$. This is motivated by the fact that the WordNet-based multi-sense vectors tend to be roughly ordered by frequency of occurrence (see analysis in Section \ref{sec:study_sense-likelihood}).

\item \textbf{Random multi-sense} evaluates using a randomly selected multi-sense vector.

\item The model \textbf{not-pretrained} is based on the approach of \citet{Kartsaklis2018Mapping}. They recently proposed a method to obtain single-sense and multi-sense vector embeddings during training (in contrast to our use of pre-trained embeddings for both). While one of their experiments also evaluates on a reverse-dictionary setting, their results are unfortunately not directly comparable, as their targets are WordNet synsets and not words. We, therefore, integrate their proposed technique into our architecture in two ways: For the model \textit{not pre-trained}, we use their equivalent version of $\vec{v}_i$. This means that we use their code for the training of the single and multi-sense embeddings as well as for the creation of $\vec{v}_i$ based on the context and the multi-sense embedding. The model \textbf{only $\boldsymbol{\es}$ pre-trained} differs from this in that we use the pre-trained single-sense embedding instead of training it from scratch.

\item The \textbf{BERT} model belongs to the class of contextual word embeddings. This approach has been rapidly become popular with works by \citet{ELMO}, \citet{GenerativePretraining}, \citeauthor{Peters2018ContextualEmbeddings} (2018b) and \citet{devlin2018bert}. Instead of using a direct mapping of words to vector representations, these approaches pre-train a neural language model on a large amount of text. The language model's internal state for each input word is then used as a corresponding word vector representation for a different task. They can be viewed as inducing word vector representations that are specific to the surrounding context. We compare against the current state-of-the-art model BERT \citep{devlin2018bert}. For this, the output of BERT's last Transformer layer is used as the sequence $(\vec{v}_1, ..., \vec{v}_n)$.
\end{itemize}

\subsection{Hyperparameters}
We follow the choices of \citet{hill2016dicteval} with an LSTM layer size of 512, a linear dense layer that maps to the size of the target vector and a batch size of 16. The input descriptions are clipped to a maximum length of 20 words and the number of senses per word is limited to 20. If a word does not exist in the multi-sense embedding, we fall back to the single-sense embedding. The pre-trained single and multi-sense word embeddings have a dimensionality of 300 and are fixed during training. For the embeddings created during training with the method of \citeauthor{Kartsaklis2018Mapping}, we experiment with the same dimensionality of 300 as well as with an embedding size of 150 (as suggested in their work). Apart from this, we follow the configuration of \citeauthor{Kartsaklis2018Mapping}\ for their components. For the contextual BERT embeddings, the authors' pre-trained, uncased model is used in the ``base'' and ``large'' variation and the pre-trained embeddings are again fixed. Since the BERT embeddings have a higher dimensionality (768 and 1024 respectively), the model architecture might underfit. We, therefore, experiment with different LSTM layer sizes up to 5,120, as well as with 2 LSTM layers and with adding a layer that transforms the embeddings to the same dimensionality of 300. For optimization, Adam \citep{kingma2014adam} is used for all models except for \textit{only $\es$ pre-trained}, which achieved better results using stochastic gradient descent with a fixed learning rate of 0.01.

\subsection{Metrics}
    
For evaluation, the vocabulary is ranked according to the cosine similarity of the produced vector $\vec{\hat{t}}$ as explained in Section \ref{sec:reverse_dictionary}. As the vocabulary, we use the union of all target words of the training, development, and test sets. Following \citet{hill2016dicteval}, we report the median rank as well as the mean accuracy @10 and @100. We also computed the mean reciprocal rank, which is a common metric in information retrieval.

\subsection{Results}
    
Table \ref{tab:result_target} shows the difference in performance between using single-sense and using multi-sense vectors as targets $\vec{t}$, as detailed in Section \ref{sec:target_vectors}. Although the number of
candidates is larger when every target word has multiple candidate target vectors, the separation of the representation of the target words into different vectors according to their senses clearly helps the model to produce a reasonable representation of the input sequence. This effect is independent of whether the input is encoded using single- or multi-sense vectors. It should be noted again that the model does not have access to the true sense during testing and that instead all possible sense vectors are used for ranking. The pre-trained, contextual BERT vectors perform very poorly as target vectors. This might be due to the larger vector size, the more complex representation or the missing or uncommon context. In fact, we found that BERT obtains only 0.009 mean reciprocal rank even if we provide it with the ground truth definitions as contexts to generate the target representations. 
    
\begin{table}
    \centering
    \begin{tabular}{cccccc}
        \toprule
        \textbf{Input Vectors} & \textbf{Target Vector} & \textbf{MR} $\downarrow$ & \textbf{Acc@10} $\uparrow$ & \textbf{Acc@100} $\uparrow$ & \textbf{MRR} $\uparrow$ \\ \midrule
        single-sense & single-sense & 535.5 & 0.115 & 0.301 & 0.067 \\ 
        single-sense & multi-sense & \textbf{135} & \textbf{0.203} & \textbf{0.458} & \textbf{0.131} \\ \midrule
        multi-sense & single-sense & 481 & 0.121 & 0.315 & 0.069 \\ 
        multi-sense & multi-sense & \textbf{107} & \textbf{0.224} & \textbf{0.490} & \textbf{0.144} \\
        \bottomrule
    \end{tabular}
    \caption{Median rank, accuracy @10 and @100 and mean reciprocal rank of single- compared to multi-sense target vectors. The first row is the model architecture proposed by \citet{hill2016dicteval}. \label{tab:result_target}}
\end{table}
    
In Table \ref{tab:result_input}, we report the results for different approaches of handling the input vectors, as introduced in Sections \ref{sec:input_vectors} and \ref{sec:multi_sense_vector_selection}. As target vectors, we use multi-sense vectors. Picking a random sense vector tends to perform slightly worse than using the single-sense vector embedding and both are outperformed by picking the first multi-sense vector of every word. This might be due to the fact that the first sense-vector tends to correspond to the most frequently occurring sense and that the representation of this sense is better in the multi-sense setting because it can focus on this meaning.
    
Using the same LSTM size of 512, the contextual BERT embeddings do not perform well. Adding a learnable linear or ReLU layer to transform them to a lower dimensionality or adding a second LSTM layer does not help either. Increasing the size of the LSTM improves performance until a certain point before it drops again. This might be due to a trade-off between the model underfitting and the learnability of the additional parameters. In the table, we report the best configuration for the "base" and "large" variation. In future work, it might also be interesting to experiment with fine-tuning the language model component of this architecture.

The model that uses the embedding training and multi-sense vector selection of \citeauthor{Kartsaklis2018Mapping} seems to struggle with building good embeddings in this setting with the 300-dimensional embeddings performing somewhat better but still not well. Providing pre-trained single-sense embeddings improves the performance considerably. Although they are not trained task-specifically, the pre-training of the single-sense embeddings on large amounts of unlabeled data seems to result in a very useful embedding space. This is consistent with other works in the literature, e.g.\ \cite{Qi2018PreTrained}. 

Our attention based multi-sense vector approach using pre-trained single- and multi-sense embeddings obtains the best results with respect to all four metrics, with the dot product similarity function performing somewhat better than cosine similarity. This shows that using pre-trained multi-sense vectors and selecting the right sense vectors can be beneficial in sequence embedding tasks. 
    
\begin{table}
    \centering
    \begin{tabular}{llccc}
        \toprule
        \textbf{Input Vectors} & \textbf{MR} $\downarrow$ & \textbf{Acc@10} $\uparrow$ & \textbf{Acc@100} $\uparrow$ & \textbf{MRR} $\uparrow$ \\ \midrule
        single-sense \citep{hill2016dicteval} & 135 & 0.203 & 0.458 & 0.131 \\ \midrule
        first multi-sense & 126 & 0.216 & 0.470 & 0.139 \\ 
        random multi-sense & 137.5 & 0.208 &    0.457 & 0.136 \\  \midrule
        not pre-trained 150 dim \citep{Kartsaklis2018Mapping} & 818 &     0.060 & 0.208 & 0.037 \\ 
        not pre-trained 300 dim \citep{Kartsaklis2018Mapping} & 574 & 0.087 & 0.260 & 0.053 \\ 
        only $\es$ pre-trained & 162 &    0.198 & 0.439 & 0.128 \\ \midrule
        BERT base LSTM 512 \citep{devlin2018bert} & 253.5 &    0.151 & 0.373 & 0.091 \\
        BERT base LSTM 4096 \citep{devlin2018bert} & 183    & 0.181 & 0.423 & 0.109 \\
        BERT large LSTM 512 \citep{devlin2018bert} & 249 & 0.156 & 0.375 & 0.093  \\
        BERT large LSTM 2048 \citep{devlin2018bert} & 220 & 0.159 &    0.391 & 0.098 \\
        \midrule
        multi-sense (cosine similarity) & 117 & 0.221 & 0.480 & 0.143 \\
        multi-sense (dot product similarity) & \textbf{107} & \textbf{0.224} & \textbf{0.490} & \textbf{0.144} \\
        \bottomrule
    \end{tabular}
    \caption{Median rank, accuracy @10 and @100 and mean reciprocal rank for the experiments with different input vectors. The multi-sense vectors are used as target vectors.\label{tab:result_input}}
\end{table}
    
\subsection{Study of Senses and Attention}
\label{sec:study_sense-likelihood}
    
In this section, we present a small study to gain more insight into the different senses occurring in the input sequences as well as into the learned attention. This is also intended as guidance for future work. For a subset of the input definitions from the training data, we manually labeled to which sense from the multi-sense embedding each word belongs. This data is made publicly available. Out of 275 words, 157 (57\%) only had one vector representation, 18 words (7\%) had a sense that was not covered by the corresponding multi-sense embedding entry, and 100 (37\%) had one sense of the multiple possible meanings provided by the multi-sense embedding. On the latter, we calculated similarly to data likelihood the sum of the probabilities that different models assign to the correct sense:
    
\begin{equation}
    L(m) = \sum_w p_m(\tau(w) \mid w),
\end{equation} 

where $m$ is the model, $w$ is a word and $\tau(w)$ is the true sense of the word. For \textit{random multi-sense}, the probability was the reciprocal of the number of senses of a word. For \textit{first multi-sense}, the probability was 1 if it was the first sense of a word in the multi-sense embedding and 0 otherwise. For \textit{attention}, we used the normalized attention $a$ of the true sense. For \textit{attention-argmax}, probability 1 was assigned to the sense that had the maximum attention. The results are given in Table \ref{tab:results_sense-likelihood}.
    
\begin{table}
    \centering
    \begin{tabular}{lc}
        \toprule
        \textbf{Model} & \textbf{L} \\ \midrule
        random multi-sense & 0.25 \\
        first multi-sense & 0.53 \\ 
        attention & 0.31 \\ 
        attention-argmax & 0.39 \\
        \bottomrule
    \end{tabular}
    \caption{Result of the analysis of the probability assigned to the true sense of multi-sense words for different models. \label{tab:results_sense-likelihood}}
\end{table}
    
As mentioned earlier, the first sense of the multi-sense embedding often reflects the dominant usage, being correct in about half of the cases. The attention approach suffers from the dilution that a soft attention entails. Due to the use of the soft-max function, all senses get at least a small amount of the probability mass. An attention mechanism that uses a more skewed probability distribution might be beneficial here. From \textit{attention-argmax}, we see that the attention method also does not always assign the largest amount of attention to the correct sense. The fact that this architecture still outperforms the others can be explained by the compositional nature of the attention mechanism. Also, some of the senses in the DeConf multi-sense embeddings tend to be very fine-grained. This means that even if not the exact sense is given the most attention, a similar sense might be. For future work, it would be interesting to improve on the context creation and sense selection component, explore options to fine-tune the embeddings as well as experiment with other multi-sense embeddings that might have a smaller number of different senses per word.
    
\section{Related Work}
\label{sec:related_work}
    
\citet{hill2016dicteval} proposed to map dictionary definitions to vectors both for the practical application of reverse dictionaries as well as to study representations of phrases and sequences. In this setting, \citet{BastosDictionaryRecursiveNetworks} experimented with recursive neural networks and additional part-of-speech information. Independently of \citeauthor{hill2016dicteval}, \citet{ScheepersCompositionality} also used dictionary definitions to evaluate ways to compose sequences of words. They studied different single-sense word embeddings and composition methods such as vector addition and recurrent neural networks. The work by \citet{Bosc2018AutoEncodingDictionary} improves word embeddings with an auto-encoder structure that goes from the target word embedding back to the definition. We consider these three works complementary to ours, as they study different single-sense architectures. 
    
In recent years, several approaches to creating multi-sense vector embeddings have been proposed. \citet{rothe-schutze:autoextend}, \citet{pilehvar2016conflated} and \citet{dasigi2017ontology} use an existing single-sense word embedding and a lexical resource to induce vectors representing different senses of a word. The latter also employ an attention-based approach for creating vectors based on the context for predicting prepositional phrase attachments. \citet{IntegrationWordSenses} use the same DeConf multi-sense embedding for integrating them in a downstream application. In contrast to our work, they require, however, a semantic network to do the disambiguation. In Sense2Vec \citep{trask2015sense2vec}, the authors create embeddings that distinguish between different meanings given the corresponding part-of-speech or named entity tag. They obtain an embedding that distinguishes e.g.\ between the location Washington and the person with the same name. The method requires the input data to be tagged with POS or NE tags. \citet{athiwilson2017} represent multiple meanings as a mixture of Gaussian distributions. The number of senses per word is fixed globally to the number of Gaussian components. \citet{RaganatoSequenceWSD} and \citet{PesaranghaderWSD} use bidirectional LSTMs to learn a mapping between words and multiple senses (not sense vectors) as a supervised sequence prediction task requiring sense-tagged text. An extensive survey on further ideas and work regarding vector representations of meaning is given by \citet{SurveyVectorMeaning}.

\citet{tang2018self} analyzed different attention mechanisms in the specific context of ambiguous words in machine translation. They limit their approach, however, to single-sense vectors and the established method of using attention over other parts of the sentence to improve the translation process.

\section{Conclusion}
    
In this work, we study the use of multi-sense vector embeddings for the reverse dictionary task. We show that single-sense embeddings such as word2vec do not adequately reflect all meanings of polysemes and homonyms and that improvements can be obtained by using multi-sense embeddings both for the target words and for the words in the input description. For the latter, we proposed a method based on attention that automatically selects the correct sense from a set of pre-trained multi-sense vectors depending on the context in an end-to-end fashion. It outperforms single-sense vectors,  multi-sense embeddings trained in a task-specific way as well as  pre-trained contextual embeddings. Our analysis of the sense selection process shows avenues for interesting future work.

\section*{Acknowledgment}

The authors would like to thank the reviewers for their helpful comments. Michael Hedderich thankfully acknowledges the support by the obtained fellowship within the FITweltweit program of the German Academic Exchange Service (DAAD). Gerard de Melo's research is in part supported by the 
Defense Advanced Research Projects Agency (DARPA) and the Army Research Office (ARO) under Contract No.\ W911NF-17-C-0098. Any opinions, findings and conclusions, or recommendations expressed in this material are those of the authors and do not necessarily reflect the views of the funding agencies.

\bibliographystyle{chicago}
\bibliography{multi-sense}
    
\end{document}